\documentclass[nohyperref]{article}
\usepackage{microtype}
\usepackage[pdftex]{graphicx}
\usepackage{subfigure}
\usepackage{booktabs} %

\usepackage[hidelinks]{hyperref}

\usepackage[accepted]{icml2023}
\usepackage{xcolor}
\usepackage{enumitem}
\usepackage{natbib}
\usepackage{hyperref}

\setlist[itemize]{leftmargin=*}
\setlist[enumerate]{leftmargin=*}

\usepackage{bm}
\usepackage{comment}  
\usepackage{graphicx}
\usepackage{url}
\usepackage{microtype}
\usepackage{multirow}
\usepackage{makecell}
\usepackage{enumitem}
\usepackage{booktabs}
\usepackage{xspace}

\usepackage[frozencache=true,cachedir=_minted-in_context]{minted} %

\newcommand{\main}{HR$\rightarrow$LR}
\newcommand{\ours}{\textbf{\texttt{SAICL}}\xspace}

\usepackage{mathtools}
\usepackage{amssymb}
\usepackage{latexsym}
\usepackage{dsfont}
\usepackage{mathrsfs}
\usepackage{amssymb}
\usepackage{amsfonts}
\usepackage{bm}
\usepackage{xspace}
\usepackage{amsthm}

\newcommand*{\argmax}{\mathop{\mathrm{argmax}}}

\newcommand{\kbf}{\mathbf{k}}

\newcommand{\qbf}{\mathbf{q}}

\newcommand{\vbf}{\mathbf{v}}

\newcommand{\xbf}{\mathbf{x}}

\newcommand{\zbf}{\mathbf{z}}

\newcommand{\Ocal}{\mathcal{O}}

\ifx\BlackBox\undefined
\newcommand{\BlackBox}{\rule{1.5ex}{1.5ex}}  %
\fi

\ifx\QED\undefined
\def\QED{~\rule[-1pt]{5pt}{5pt}\par\medskip}
\fi

\ifx\proof\undefined
\newenvironment{proof}{\par\noindent{\em Proof:\ }}{\hfill\BlackBox\\[.0mm]}
\fi

\ifx\theorem\undefined
\newtheorem{theorem}{Theorem}[section]
\fi

\ifx\example\undefined
\newtheorem{example}{Example}[section]
\fi

\ifx\property\undefined
\newtheorem{property}{Property}
\fi

\ifx\lemma\undefined
\newtheorem{lemma}{Lemma}[section]
\fi

\ifx\proposition\undefined
\newtheorem{proposition}{Proposition}[section]
\fi

\ifx\remark\undefined
\newtheorem{remark}{Remark}
\fi

\ifx\corollary\undefined
\newtheorem{corollary}{Corollary}[section]
\fi

\ifx\definition\undefined
\newtheorem{definition}{Definition}[section]
\fi

\ifx\conjecture\undefined
\newtheorem{conjecture}{Conjecture}[section]
\fi

\ifx\axiom\undefined
\newtheorem{axiom}[theorem]{Axiom}
\fi

\ifx\claim\undefined
\newtheorem{claim}[theorem]{Claim}
\fi

\ifx\assumption\undefined
\newtheorem{assumption}{Assumption}
\fi

\ifx\problem\undefined
\newtheorem{problem}{Problem}[section]
\fi

\ifx\fact\undefined
\newtheorem{fact}{Fact}
\fi

\newcommand{\rbr}[1]{\left(#1\right)}

\newcommand{\cbr}[1]{\left\{#1\right\}}

\newcommand{\revise}[1]{#1}%

\icmltitlerunning{\revise{Scaling In-Context Demonstrations with Structured Attention}}
\ifdefined\usebigfont

\usepackage{times}
\usepackage[fontsize=13pt]{scrextend}
\makeatletter
\@ifpackageloaded{geometry}{\AtBeginDocument{\newgeometry{letterpaper,left=1.56in,right=1.56in,top=1.71in,bottom=1.77in}}}{\usepackage[letterpaper,left=1.56in,right=1.56in,top=1.71in,bottom=1.77in]{geometry}}
\makeatother
\else
\fi

\begin{document}
\ifdefined\usebigfont
\onecolumn
\icmltitle{How to Make Your Language Model\\a Better In-context Learner?}
\else
\twocolumn[
\icmltitle{Scaling In-Context Demonstrations with Structured Attention}

\icmlsetsymbol{equal}{*}

\begin{icmlauthorlist}
    \icmlauthor{Tianle Cai}{equal,ppp}
    \icmlauthor{Kaixuan Huang}{equal,ppp} 
    \icmlauthor{Jason D.~Lee}{ppp}
    \icmlauthor{Mengdi Wang}{ppp} 
\end{icmlauthorlist}
\icmlaffiliation{ppp}{Princeton University}

\icmlkeywords{In-context Learning, Efficient Transformer}
\vskip 0.3in]
\fi
\printAffiliationsAndNotice{\icmlEqualContribution}

\begin{abstract}
    The recent surge of large language models (LLMs) highlights their ability to perform in-context learning, i.e., ``learning'' to perform a task from a few demonstrations in the context without any parameter updates.
However, their capabilities of in-context learning are limited by the model architecture: 1) the use of demonstrations is constrained by a maximum sentence length due to positional embeddings; 
2) the quadratic complexity of attention hinders users from using more demonstrations efficiently; 3) LLMs are shown to be sensitive to the order of demonstrations. 
In this work, we tackle these challenges by proposing a better architectural design for in-context learning.
We propose \ours (\textbf{S}tructured \textbf{A}ttention for \textbf{I}n-\textbf{C}ontext \textbf{L}earning), which replaces the full-attention by a structured attention mechanism designed for in-context learning, and removes unnecessary dependencies between individual demonstrations, while making the model invariant to the permutation of demonstrations.
We evaluate {\ours} in a meta-training framework and show that {\ours} achieves comparable or better performance than full attention while obtaining up to 3.4x inference speed-up. {\ours} also consistently outperforms a strong Fusion-in-Decoder (FiD) baseline which processes each demonstration independently. Finally, thanks to its linear nature, we demonstrate that {\ours} can easily scale to hundreds of demonstrations with continuous performance gains with scaling.

\end{abstract}

\section{Introduction}\label{sec:intro}
    Large language models (LLMs) have recently made notable advances with superior performance in downstream tasks~\citep{brown2020language,chowdhery2022palm,zhang2022opt,hoffmann2022training}. One emergent property of LLMs is their ability to learn \emph{in-context}~\citep{brown2020language}, i.e., with a few task demonstrations provided in the prompt, LLMs are readily adapted to make accurate predictions without any parameter updates.  The process of in-context learning only requires a single forward pass, making it an efficient and flexible alternative to fine-tuning for many real-world problems.

There are two major limitations of existing LLMs' in-context learning. \emph{First}, it is expensive or infeasible to support a large number of demonstrations (typical choices ranging from 5~\citep{chung2022scaling} to 32~\citep{brown2020language}) due to the quadratic complexity of their attention mechanisms and the maximal sentence length constraints of the inputs. Therefore, whether scaling up the number of demonstrations will improve the performance of in-context learning remains unexplored. \emph{Second}, the demonstrations are sequentially concatenated in the prompt, and recent evidence shows that performance is sensitive to the order of the demonstrations~\citep{lu2021fantastically,zhao2021calibrate}. This artifact creates additional overhead and design for selecting a better order of demonstrations~\cite{lu2021fantastically}.

In this paper, we seek novel solutions to address the above challenges by \emph{removing unnecessary dependencies between the demonstrations}, and \emph{making the model invariant to their permutations}.
This problem has yet to be comprehensively studied in the literature, while some existing ideas can be adapted. In particular, we find that the Fusion-in-Decoder (FiD) model~\citep{izacard2021leveraging}\footnote{A concurrent work~\cite{ye2022investigating} also experiments with FiD for in-context learning. See more discussion in related work.}---originally proposed for open-domain question answering---is a strong baseline for in-context learning. In FiD, demonstration-input pairs are \emph{independently} encoded and then concatenated before feeding into the decoder\footnote{FiD relies on the encoder-decoder Transformer architecture like T5~\citep{2020t5}.}. Independent encoding of demonstrations makes FiD scale \emph{linearly} with the number of demonstrations, at the expense of fusing demonstrations only at the decoder stage. Another simple approach is \emph{ensemble}, which averages the predictions based on individual demonstrations~\citep{min2021noisy}, and the dependencies between the demonstrations are completely ignored. Although both FiD and ensemble are already efficient and permutation-invariant, we find that their performance is prone to saturate empirically as more demonstrations are used (Figure~\ref{fig:avn}, \ref{fig:a1}), and they often underperform standard concatenation-based baselines (Table~\ref{tab:main-result}, Appendix~\ref{sec:ensemble}).  Therefore, how to strike the balance between efficient encoding and retaining necessary dependencies between demonstrations is the key research question to explore.

\begin{figure}[tbp]
   \centering
    \includegraphics[width=\linewidth]{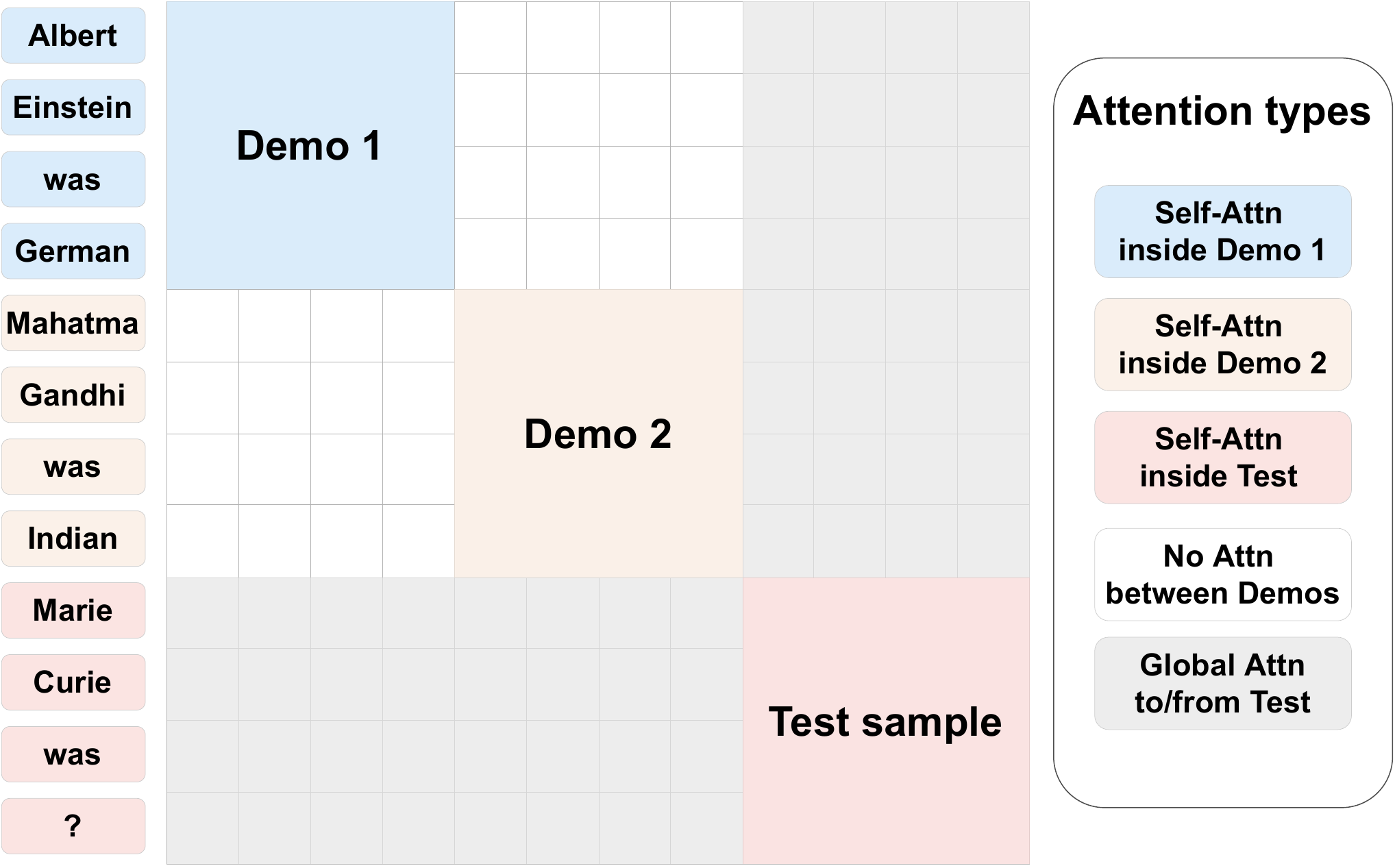}
    \caption{An illustration of \ours with two demonstrations. \ours is a structured attention mechanism designed for in-context learning that can directly replace the self-attention layers in Transformer models. By removing the attention across different demonstrations while keeping the test input to attend to all tokens globally, we reduce the redundancy in computation but retain necessary dependencies between different demonstrations. \ours, therefore, 1) enjoys linear computational and memory complexity; 2) has no limitation on the number of demonstrations used; 3) is invariant to the permutation of the demonstrations. \texttt{Demo} $i$ stands for the $i$-th demonstration, \texttt{test} stands for the test input. %
    }
    \label{fig:cfa}
    
\end{figure}

Motivated by the recent development of sparse Transformers~\citep{child2019generating, beltagy2020longformer, zaheer2020big}, we propose \ours (\textbf{S}tructured \textbf{A}ttention for \textbf{I}n-\textbf{C}ontext \textbf{L}earning) as a replacement for the full-attention mechanism. Unlike existing sparse attentions that approximate the capability of full attention to process any input sequence, the design of \ours is \emph{tightly coupled with the structure of in-context learning}. As shown in Figure~\ref{fig:cfa}, we only allow the tokens of each demonstration to attend to themselves and the tokens of the test input, while the attentions of the tokens of the test input remain unchanged. In this way, the information from each demonstration is fused through the \emph{global attention} of the test input in each attention layer and then sent back to all demonstrations in the next attention layer. Therefore, each demonstration is able to utilize the information from all other demonstrations while the complexity of the attention module is still \emph{linear} to the number of demonstrations. Additionally, the positional encoding is only computed inside each demonstration, making the model invariant to the permutation of the demonstrations.

As \ours requires modifications to the model architecture, we evaluate it in a meta-training framework~\citep{min2021metaicl} for in-context learning. 
We use T5~\citep{2020t5} as our base model and compare \ours with standard T5 models with full attention and the FiD model {(See Section~\ref{sec:cf_attention} for a discussion about the choice of using T5)}. Specifically, we meta-train all models with the prompt being the same format of in-context learning using a set of source tasks and evaluate their performance when transferred to unseen target tasks. 
Our experiments (Section~\ref{sec:cf_dependency}) indicate that \ours often achieves better performance than  FiD ($18$ wins out of $28$ settings) and is able to match or even beat ($13$ wins out of $28$ settings) the full attention baseline. Furthermore, we demonstrate that \ours can scale to hundreds of demonstrations, and improve its performance consistently with more demonstrations (Figure~\ref{fig:avn}). Thanks to its linear complexity, {\ours} also achieves significant speed-ups during inference---up to 3.4x for 64 demonstrations and 6.3x for 128 demonstrations.

\section{Related Work}\label{sec:related_work}
    \paragraph{In-context learning.} 
Large language models can perform in-context learning~\citep{brown2020language}, where LLMs are adapted to new tasks by conditioning them on a few demonstrations. 
Although in-context learning is shown mainly with decoder-only models such as GPTs\citep{radford2019language,brown2020language}, several recent works also experiment with encoder-decoder models such as T5 \citep{patel2022bidirectional, chung2022scaling, tay2022unifying}. 

Many recent works aim to understand how in-context learning works through empirical and theoretical investigation~\citep{xie2021explanation, min2022rethinking,akyurek2022learning, garg2022can, von2022transformers, dai2022can, chan2022data,lyu2022z,qiao2022reasoning,li2023transformers}. Some other studies focus on improving the performance of in-context learning through calibration~\citep{zhao2021calibrate, holtzman-etal-2021-surface,min2021noisy}, self-supervised training \citep{chen-etal-2022-improving}, and meta-learning \citep{min2021metaicl, chung2022scaling}. Other directions include retrieving demonstrations that are semantically similar to the test input \citep{liu2022makes, rubin2021learning}, or using chain-of-thought prompts to enhance reasoning~\citep{wei2022chain,kojima2022large,wang2022self,wang2022towards}.

One notorious property of in-context learning is its sensitivity to the order of demonstrations. \citet{zhao2021calibrate} show that LLMs are prone to outputting the same label as the last demonstration (aka recency bias). \citet{lu2021fantastically} find that under some permutation of the demonstrations, the performance is close to random guesses, and this phenomenon exists across various models. As a remedy, they propose to exhaustively search for all possible prompt orders and use entropy statistics to select the best order. The permutation sensitivity of in-context learning incurs a computational overhead to search for a good prompt order. This motivates us to address the order-sensitivity issue of in-context learning through a better model design. 

\paragraph{Efficient Transformers.} Despite the success of the Transformer architecture~\citep{vaswani2017attention} in a wide range of machine learning applications, its computation and memory complexity scale quadratically with the input length. This makes it intractable for modeling long-range contexts. Therefore, a variety of efficient Transformer models have been proposed to mitigate this issue~\citep{child2019generating,beltagy2020longformer,wang2020linformer,kitaev2019reformer,katharopoulos2020transformers,zaheer2020big,qiu2020blockwise,tay2020synthesizer,peng2021random,qin2021cosformer,zhou2021informer}. We refer the readers to \citet{tay2022efficient} for a comprehensive survey. {In this paper, we draw inspiration from the design of sparse attention mechanisms in efficient Transformers~\citep{beltagy2020longformer,zaheer2020big}. However, instead of using sparse attention to approximate the ability of full attention to process any sequences, we exploit the inherent structure of in-context learning (test input follows a number of demonstrations), and design a structured attention mechanism tailored for the problem.

\paragraph{Comparison with concurrent works.} Concurrent with our work, there are several works~\citep{hao2022structured, ratner2022parallel, ye2022investigating} exploring architectural designs to improve the in-context learning ability of LLMs. Among them, \citet{hao2022structured, ratner2022parallel} focus on tackling the limitation of context length by independently processing several prompts \emph{within the max length constraint} and then combining them. This idea is similar to our adaptation of FiD while \citet{hao2022structured, ratner2022parallel} use \emph{fixed schemes} (e.g., average) for fusing parallel prompts so that their methods can be used directly on top of pre-trained models without further fine-tuning. Compared to their methods, we, in addition, seek to address the efficiency and instability issues while exploring better encoding to retain the dependencies between demonstrations. These goals cannot be directly achieved by processing several grouped prompts in parallel. \citet{ye2022investigating} investigate different ways of fusing demonstrations, and they also observe that adapting FiD to in-context learning is effective (usually better than the ensemble-based method and more efficient than the concatenation-based method), which is aligned with our findings (Figure~\ref{fig:a1} and Section~\ref{sec:cf_dependency}). %

Due to length limitation, we refer the readers to Appendix~\ref{sec:more_related} for more related works on (few-shot) fine-tuning.

\section{\revise{Efficient In-Context Learning with SAICL}}\label{sec:method}%
\subsection{Desiderata of In-Context Learning Models}\label{sec:desiderata}
    For in-context learning, assume we are given $k$ demonstrations $(\xbf_i, y_i)_{i=1}^k$, we construct the prompt as the concatenation of all the demonstration pairs and the test input: $\xbf_\mathrm{prompt} = (\xbf_1, y_1, \xbf_2, y_2, \dots, \xbf_k, y_k, \xbf_\mathrm{test})$.  When we  evaluate the model on a fresh input $\xbf_\mathrm{test}$ to predict from a set of candidate answers $Y_\mathrm{test} = \{y_\mathrm{test}^{(1)}, \dots, y_\mathrm{test}^{(m)} \}$, we condition the language model on $\xbf_\mathrm{prompt}$ and query the probabilities of its continuations to inference about the answer $p(y_\mathrm{test}\mid\xbf_\mathrm{prompt})$ for each $y_\mathrm{test} \in Y_\mathrm{test}$. \footnote{We use the \emph{direct} method for demonstration in this section. See Section~\ref{sec:settings} for the comparison between the \emph{direct} method and the \emph{channel} method.} Afterwards, the answer for $\xbf_\mathrm{test}$ is $\argmax_{y_\mathrm{test}} p(y_\mathrm{test}\mid \xbf_\mathrm{prompt})$. For encoder-decoder models such as T5, the $\xbf_\mathrm{prompt}$ is fed into the encoder, whose embedding is then computed as key-value pairs for the cross-attention modules of the decoder to perform auto-regressive decoding. 

We state the high-level design goals for an ideal in-context learner. 
1) \textbf{Extensibility.} The model should be capable of using many demonstrations to boost performance without any \emph{hard-coded constraints} on the number of demonstrations caused by the model design, e.g., the max length constraints due to the limited number of positional encodings.  
2) \textbf{Efficiency.} The model should be able to \emph{efficiently} scale to a large number (e.g., hundreds) of demonstrations. Ideally, the computational complexity should only have a linear dependency on the number of demonstrations. 
3) \textbf{Invariance.} The model should be invariant to the permutation of demonstrations, which saves extra computation on searching over the \emph{combinatorial space of possible permutations} and preserving the natural symmetry of demonstrations.

\subsection{\ours}\label{sec:cf_attention}
    We propose \textbf{S}tructured \textbf{A}ttention for \textbf{I}n-\textbf{C}ontext \textbf{L}earning (\ours, pronounced as \emph{cycle}) that replaces the full bidirectional attention in the \emph{encoder} of encoder-decoder models such as T5. 
In \ours, we exploit the special structure of the prompts of in-context learning. We observe that the computational burden comes from the dense attention between \emph{all pairs of demonstrations}. Intuitively, the understanding of each demonstration $(\xbf_i, y_i)$ only \emph{loosely} depends on the understanding of all other demonstrations. Therefore, in \ours, we remove the attentions across different demonstrations as shown in Figure~\ref{fig:cfa}. Meanwhile, we keep the attention from $\xbf_\mathrm{test}$ global to make $\xbf_\mathrm{test}$ an aggregator of information from different demonstrations. Concretely, the tokens of the $i$-th demonstration $(\xbf_i,y_i)$ can only attend to 1) tokens within the same demonstration and 2) the test tokens, while the test tokens $\xbf_\mathrm{test}$ can attend to all the tokens. This way, the test input tokens can fuse information from all demonstrations and send information back to each demonstration in the next layer, which implicitly makes each demonstration fully utilize information from all other demonstration exemplars.

For simplicity of illustration, we assume each sample only has \emph{one token} and ignore the relative positional encodings (for a comprehensive description of \ours, please refer to the pseudo-code in Appendix~\ref{sec:python_pseudo_code}). Let $\cbr{(\qbf_i, \kbf_i, \vbf_i)}_{i=1}^k$ and $(\qbf_{\mathrm{test}}, \kbf_{\mathrm{test}}, \vbf_{\mathrm{test}})$ be the (query, key, value) triplets of demonstrations and test sample for calculating the attention, respectively. Then the output of \ours at position $i\in[1, k]$ is calculated as:
\begin{align*}
    \zbf_i = \frac{\exp\rbr{\qbf_i^\top \kbf_i}\vbf_i + \exp\rbr{\qbf_i^\top \kbf_{\mathrm{test}}} \vbf_{\mathrm{test}}}{\exp\rbr{\qbf_i^\top \kbf_i} + \exp\rbr{\qbf_i^\top \kbf_{\mathrm{test}}}};
\end{align*}
and the output at the test sample position is
\begin{align*}
    \zbf_{\mathrm{test}} = \frac{\sum_{i=1}^k \exp\rbr{\qbf_{\mathrm{test}}^\top\kbf_i}\vbf_i+\exp\rbr{\qbf_{\mathrm{test}}^\top\kbf_{\mathrm{test}}}\vbf_{\mathrm{test}}}{\sum_{i=1}^k \exp\rbr{\qbf_{\mathrm{test}}^\top\kbf_i}+\exp\rbr{\qbf_{\mathrm{test}}^\top\kbf_{\mathrm{test}}}}
\end{align*}
\paragraph{Efficiency.} This sparse structure of \ours then makes the computational complexity and the memory consumption only \emph{linearly} depend on the number of demonstrations. \revise{Concretely, assume there are $k$ demonstrations, and each demonstration and test sample has a max length $L$, then the computational and memory complexity of full attention will be $\Ocal(k^2L^2)$ while the complexity of \ours will be $\Ocal(kL^2)$.} As shown in Figure~\ref{fig:fs}, we create fake inputs where the number of tokens in each demonstration is fixed to 64 and 128 for evaluating the scalability of inference time of \ours and full attention. The reported inference times are averaged across ten runs tested on RTX A6000 GPUs. We observe that \ours scales much better than the full attention, yet the advantage of \ours is less significant when the number of demonstrations is small. We hypothesize that this is mainly due to the implementation overhead of \ours, which may be further improved. 

\paragraph{Extensibility and permutation invariance.} Unlike the decoder-only GPT models, the T5 encoder only relies on the relative positional encodings to determine the order of the tokens. Specifically, instead of adding an absolute positional encoding to each token embedding, T5 adds a bias term to the pre-softmax value of the attention heads, which only depends on the difference between the query position and the key position. \revise{In \ours, we only keep the relative positional encodings inside the tokens of each demonstration or test sample. Therefore, the max length constraint only limits the length of each sample, which is usually short in practice.} Also, in T5, the order of demonstrations is distinguished solely via the relative positional encodings in the attentions \emph{across} difference demonstrations. Thus, after removing these attentions, our \ours automatically achieves the permutation invariance goal. 

\revise{\paragraph{Discussion on the choice of encoder-decoder model.} We build \ours and conduct experiments on top of the T5 model, and many ideas benefit from the T5 architecture: 1) \textbf{More flexible information exchange.} The bidirectional attention mechanism enables free information exchange between different demonstrations. Specifically, in \ours, the information can be passed through the global attention by the test sample. Also, the encoder-decoder separation enables parallel encoding in FiD. 2) \textbf{Easier to keep the symmetry.} It is more natural to use bidirectional attention for encoding demonstrations because there is no canonical order of demonstrations, but the causal mask in decoder-only models induces an order. Also, as mentioned before, the relative positional encoding in T5 enables the permutation invariance in \ours. Yet, we believe that it is easy to adapt our methods to a decoder-only model with modifications on the positional encoding and the attention mask.}

\begin{figure}[tbp]
    \includegraphics[width=\linewidth]{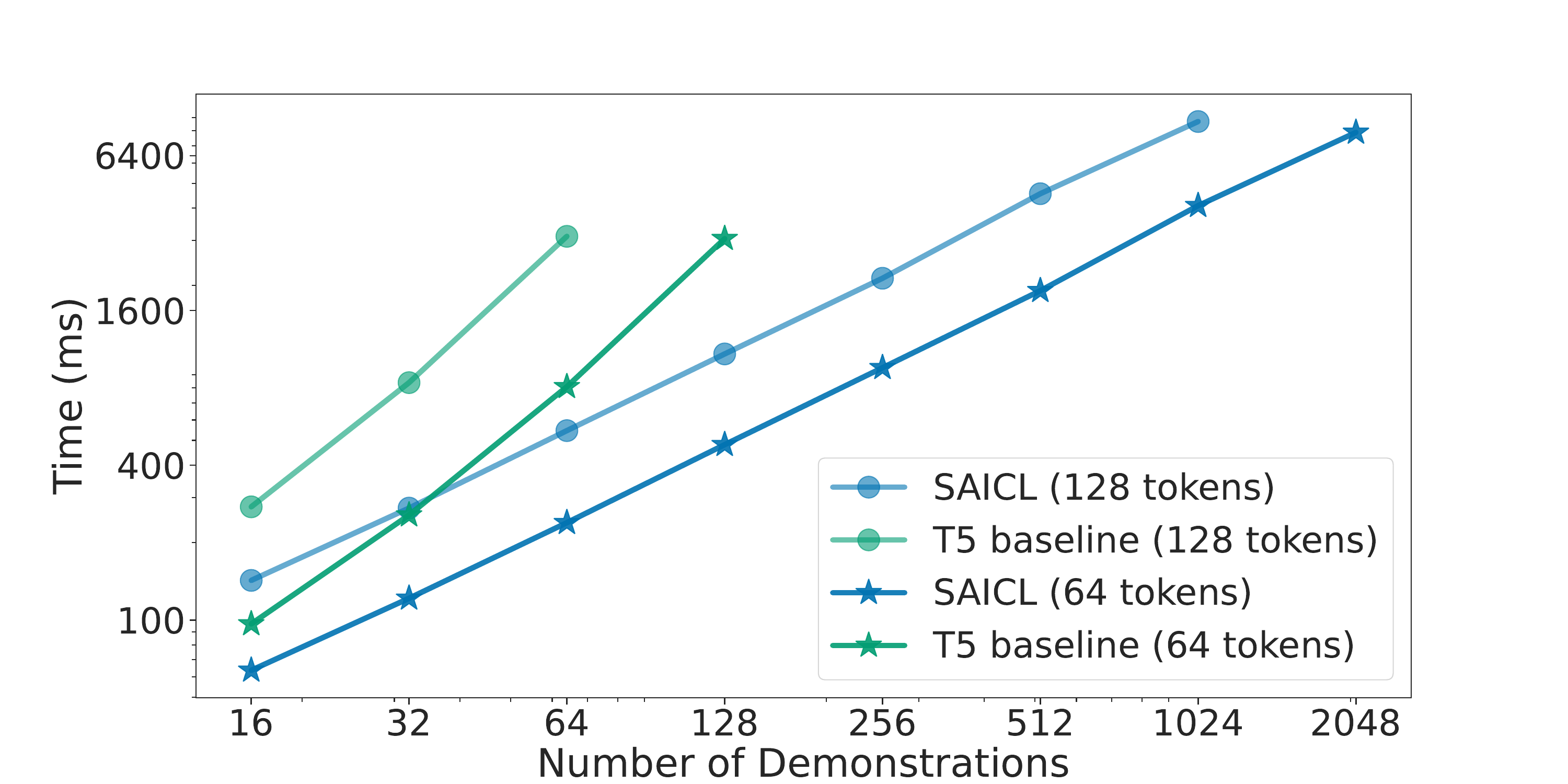}
    \caption{The inference times (ms) v.s. different numbers of demonstrations. \revise{The inference time of \ours consistently scales better than the full attention baseline with different per sample length $L\in\{64, 128\}$. By design, the computational complexity of \ours scales linearly to the number of demonstrations $k$ while the full attention scales quadratically ($\Ocal(kL^2)$ v.s. $\Ocal(k^2L^2)$). Also note that the T5 baseline runs out of memory quickly (64 shots), even on an A6000 with 50 GB of memory.}} %
    \label{fig:fs}
\end{figure}

\subsection{Meta-Training}\label{sec:meta_train}
    We use meta-training to fine-tune the models following Meta In-Context Learning (MetaICL)~\citep{min2021metaicl}.
 The meta-training objective is 
\begin{align*} 
    \max_\theta  \hat{\mathbb{E}} \log p_{\theta}(y_\mathrm{test}\mid \xbf_1, y_1, \dots, \xbf_k, y_k, \xbf_\mathrm{test}),
\end{align*}
where the model is trained using a mixture of source tasks and the prompt has the same format of in-context learning, i.e., the concatenation of the demonstrations and the test input. 
During training, we first sample a source task and then randomly draw $(\xbf_i,y_i)$'s and $(\xbf_\mathrm{test},y_\mathrm{test})$  from its training dataset. %

We use the same datasets, evaluation protocols, and prompt template designs as MetaICL. Concretely, we use 142 tasks from CrossFit~\citep{ye2021crossfit} and UnifiedQA~\citep{khashabi2020unifiedqa}, which include text classification, question answering (QA), natural language inference (NLI), and paraphrase detection. They are divided into seven transfer settings, each consisting of a non-overlapping pair of source and target tasks. A subset of the target tasks is from completely different domains from the training tasks, e.g., medical, financial, and climate. Following \citet{min2021metaicl}, we highlight these tasks as unseen domain tasks, where the gains of in-context learning over the multi-task fine-tuning setting are more significant.

At test time, we evaluate the meta-trained model on the unseen target tasks with $k$-shot demonstrations and average the results across $5$ different random sets of the demonstrations. Macro-F1 and accuracy are used as evaluation metrics for classification tasks and non-classification tasks, respectively. In order to investigate the scaling behavior of in-context learning, for each transfer case and each method, we meta-train two models with different training-time numbers of demonstrations: train $k=16$ and train $k=64$. Then, we use the corresponding fine-tuned model at test time and evaluate it under test $k=16$ and test $k=64$ settings. In ablation studies, we use the high-resource to low-resource transfer setting (HR$\to$LR) following \citet{min2021metaicl}. This setting is the most representative one because its source and target both cover all the task types.

\section{Experimental Setup}\label{sec:exp_setup}
\subsection{Settings}\label{sec:settings}
    \paragraph{Base models.} Different from MetaICL \citep{min2021metaicl}, we choose T5-LM-adapt large (770M parameters)\footnote{\url{https://huggingface.co/google/t5-large-lm-adapt}} as our base model, which is additionally fine-tuned from T5.1.1 using LM objective\footnote{This LM-adapted version of T5 is suggested to have better in-context learning ability than the original ones~\citep{sanh2021multitask}.}. Please see Section~\ref{sec:cf_attention} for a discussion about the choice of using T5.%

\paragraph{Direct v.s. channel.} There are two mainstream methods to perform in-context learning: direct method and channel method \citep{min2021noisy}. For \emph{direct} method, the LM is prompted with $(\xbf_1, y_1, \dots, \xbf_k, y_k, \xbf_\mathrm{test})$ to predict the probability of the answer $y_\mathrm{test}$. %
For \emph{channel} method, the LM is prompted with $(y_1, \xbf_1, \dots, y_k, \xbf_k, y_\mathrm{test})$ to predict the probability of the test input $\xbf_\mathrm{test}$.
As suggested by \citet{min2021noisy, min2021metaicl}, the channel method usually performs better than the direct method thanks to better calibration, so we choose to use the \emph{channel} method in \emph{all} our experiments. %

For detailed data processing and optimization settings, please see Appendix~\ref{sec:detail_setting}.
\subsection{Baselines}\label{sec:baselines}
    We compare \ours with the following baseline methods. 

\paragraph{T5 baseline.} We meta-train the LM-adapted T5-large model without modifying its attention mechanism using the channel method. In this case, the computational complexity is $\mathcal{O}(k^2)$.

\paragraph{Fusion-in-Decoder (FiD) \citep{izacard2021leveraging}.} We adapt the FiD to in-context learning by independently passing each $\xbf_{\mathrm{prompt},i} = (y_i, \xbf_i, y_\mathrm{test})$ through the standard T5 encoder and concatenate all their outputs, and then feed the concatenation into the cross-attention layers of the T5 decoder to calculate the conditional probability of $x_\mathrm{test}$. The computational complexity is $\mathcal{O}(k)$ for FiD.

\paragraph{MetaICL baselines.}

We report four baseline results from the original MetaICL paper~\citep{min2021metaicl} where \emph{GPT-2 large models} are used: 
\begin{itemize}[itemsep=1pt, parsep=0pt, topsep=0pt]
    \item Multi-task 0-shot: Meta-trained without any demonstration using \emph{direct} method. 
    \item Channel Multi-task 0-shot: Meta-trained without any demonstration using \emph{channel} method.  
    \item MetaICL: Meta-trained with $k=16$ demonstrations using \emph{direct} method.   
    \item Channel MetaICL: Meta-trained with $k=16$ demonstrations using \emph{channel} method.   
\end{itemize}

\section{Experiments}\label{sec:experiments}
\subsection{Main Results}\label{sec:cf_dependency}

\begin{table*}[h]
    \centering \footnotesize
    \begin{tabular}{
        l @{\hspace{2em}}
        cccccccc
        }
        \toprule
            Method & 
            \makecell[c]{Complexity} &
            \main &
            \makecell[c]{Class \\ $\rightarrow$Class} & \makecell[c]{non-Class \\ $\rightarrow$Class} &
            \makecell[c]{QA \\ $\rightarrow$QA} &
            \makecell[c]{non-QA \\ $\rightarrow$QA} &
            \makecell[c]{non-NLI \\ $\rightarrow$NLI} &
            \makecell[c]{non-Para \\ $\rightarrow$Para} \\
        \midrule
            \multicolumn{9}{c}{\em All target tasks} \\
            Multi-task 0-shot\footnotemark[1] & \multirow{2}{*}{$\Ocal(1)$} & 35.6 & 37.3 & 36.8 & 45.7 & 36.0 & 40.7 & 30.6 \\
            Channel Multi-task 0-shot\footnotemark[1] &  & 38.8 & 40.9 & 42.2 & 42.1 & 36.4 & 36.8 & 35.1 \\
        \cmidrule{1-9}
            MetaICL\footnotemark[1] & \multirow{4}{*}{$\Ocal(k^2)$} & 43.3 & 43.4 & 38.1 & {46.0} & 38.5 & 49.0 & 33.1 \\
            Channel MetaICL\footnotemark[1] & & {49.1} & {50.7} & {50.6} & 44.9 & {41.9} & \textbf{54.6} & {52.2} \\
            T5 baseline &  & 50.8 & 51.9 & 54.7 & 46.2 & 44.5 & 45.7& 51.3 \\
            T5 baseline ($64$-shot) & & 52.7  & 57.2 & \textbf{58.9} & 46.4 & 44.6 & 48.1 & 55.6 \\
        \cmidrule{1-9}
            FiD & \multirow{4}{*}{$\Ocal(k)$} & 49.0 & 50.1 & 50.7 & 45.7 & 44.2 & 48.1 & 51.7 \\
            \ours &  & 49.8 & 51.0 & 50.9 & 45.7 & 43.8 & 44.8 & 51.8 \\
            FiD ($64$-shot) & & 50.7 & 57.6 & 48.6 & \textbf{46.6} & \textbf{45.3} & 53.7 & 57.2 \\
            \ours ($64$-shot) & & \textbf{53.6} & \textbf{58.3} & {53.3} & 46.3 & 44.3 & 48.9 & \textbf{57.8} \\
        \toprule
            \multicolumn{9}{c}{\em Target tasks in unseen domains} \\
            Multi-task 0-shot\footnotemark[1] & \multirow{2}{*}{$\Ocal(1)$} & 35.4 & 28.0 & 28.6 & \textbf{71.2} & 40.3 & 33.5 & 35.0 \\
            Channel Multi-task 0-shot\footnotemark[1] && 36.3 & 31.1 & 34.3 & 54.4 & 39.4 & 50.8 & 34.1 \\
        \cmidrule{1-9}
            MetaICL\footnotemark[1] & \multirow{4}{*}{$\Ocal(k^2)$}& 35.3 & 32.3 & 28.1 & 69.9 & {48.3} & \textbf{80.1} & 34.0 \\
            Channel MetaICL\footnotemark[1] & & {47.7} & {41.9} & {48.0} & 57.9 & 47.2 & 62.0 & {51.0} \\
            T5 baseline & & 53.2 & 50.1 & 50.5 & 50.8 & 50.8 & 60.0 & 47.4\\
            T5 baseline ($64$-shot) & & 54.7 & 54.6 & \textbf{54.0} & 53.3 & 46.1 & 62.3 & 53.8 \\
        \cmidrule{1-9}
            FiD &\multirow{4}{*}{$\Ocal(k)$} & 49.8 & 47.3 & 47.3 & 51.8 & \textbf{53.3} & 59.8 & 50.9 \\
            \ours & & 49.0 & 52.8 & 47.9 & 52.8 & 50.4 & 57.5 & 53.8 \\
            FiD ($64$-shot) & & 55.3 & 55.1 & 40.7 & {53.3} & 48.3 & {67.6} & 58.3 \\
            \ours ($64$-shot) & & \textbf{57.4} & \textbf{56.8} & {48.9} & 52.9 & 49.4 & 63.4 & \textbf{60.8} \\
        \bottomrule
    \end{tabular}%
    \caption{Main results. \emph{We use 16 shots by default.} For the same number of demonstrations, we see that \ours performs better than FiD in 18 out of 28 settings, and is able to achieve comparable performance as the T5 baseline. Scaling up the number of demonstrations $k$ further boosts the performance for T5 baseline, FiD, and \ours. $^1$The results are copied from \citet{min2021metaicl}, where the base model used for these 4 baselines is \emph{GPT-2 large}. Our result indicates that the T5 baseline performs better than GPT-2. \textbf{Bold} indicates the best result in each column. The results are averaged across 5 different samples of demonstrations. %
    }\label{tab:main-result}
\end{table*}

We present the results in Table~\ref{tab:main-result} and summarize the findings below.

\paragraph{Our baselines (T5 and FiD) are strong.} Although T5 large is roughly the same size as GPT-2 large, our experimental results indicate that T5 baselines generally perform better than GPT-2 in the Meta-ICL setting (9 wins out of 14 settings). We hypothesize that the benefits partially come from the bidirectional attention of the T5 encoder, which, compared to decoder-only models such as GPT, allows each demonstration to utilize the information from all other demonstrations. For FiD, our experiments show that it can be seamlessly adapted to in-context learning and usually match the performance of full attention without any fusion of demonstrations in the encoder.

\paragraph{\revise{\ours enables sufficient fusion among demonstrations.}}
With the same number of demonstrations, we observe that \ours usually performs better than FiD (18 wins out of 28 settings) while maintaining the same linear complexity. Analogously, we can view our method as ``fusion in encoder'', which has more flexibility for exchanging information between demonstrations. Compared to the full attention, \ours can usually match or even beat its performance while being much more efficient. 
\revise{These results indicate that \ours enables sufficient fusion among the demonstrations as the full attention, while completely disabling information exchange in the encoder (FiD) will lead to worse performance.}

\paragraph{Scaling up the number of demonstrations $k$ boosts the performance.} Our results show that increasing $k$ from $16$ to $64$ can significantly improve the performance for all models (by $\sim 3\%$ in average). This observation is seemingly in contrast with the finding in the MetaICL paper~\citep{min2021metaicl}, where the authors show the performance improvement tends to saturate after $32$ demonstrations. We hypothesize that the difference is because the $1024$ length limitation of the GPT-2 model used in \citet{min2021metaicl} constrains the effective context. This finding suggests that the potential of boosting performance with more demonstrations might be underestimated because of inappropriate architectural design.

\begin{figure}[h]
    \includegraphics[width=\linewidth]{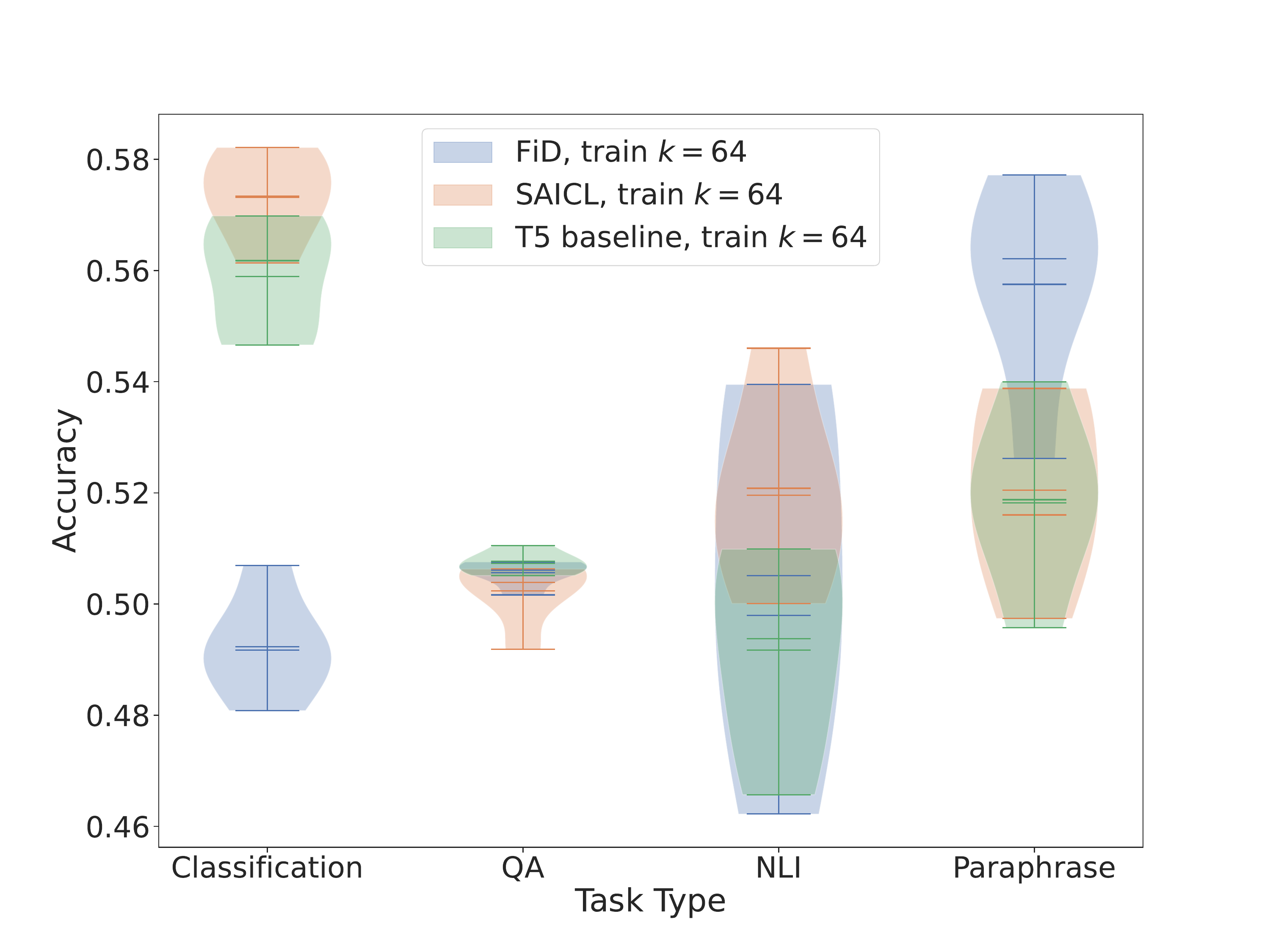}
    \caption{Comparison of \ours and FiD across task types. We group the performance of \ours and FiD in terms of task type in the {HR$\to$ LR} setting. The plot shows that the benefits of \ours are more pronounced on classification tasks. We hypothesize that classification tasks rely more on the input format and label space, which can be extracted by comparing different demonstrations, and thus, more dependencies are needed.
    }
    \label{fig:task_type}
\end{figure}

\paragraph{Different behaviors on various types of tasks.} We notice that the methods we test behave differently across different tasks, and no one can rule over all tasks. This is because solving different tasks requires different forms of information. For example, to solve classification tasks with in-context learning, it is crucial to determine the label space and the format by which a classification problem is converted to natural language.
\revise{ In Figure~\ref{fig:task_type}, we inspect the behaviors of \ours and FiD on different types of tasks under the \emph{test} domain in the HR$\to$LR setting, where all types of tasks are covered by both training and test domains.}
One observation is that the benefits of \ours are more pronounced on classification tasks. We hypothesize that this is because classification tasks require more subtle knowledge about the format and label space, which benefits from the \revise{ fusion of demonstrations enabled by \ours in the encoder}. %
This observation also coincides with the finding in \citet{min2022rethinking} that in-context learning relies on the label space, input format, and distribution of input text to work.

\paragraph{Discussion on the model size.} Due to computational limitations, we use the T5 large model (770M) in this paper. It is an interesting future work to extend our methods to larger models. We notice that in the concurrent work \citet{ye2022investigating} (Figure 2), the authors already find that the dominance of FiD compared to ensemble methods is much more significant when the model size increases from T5 large to T5 XL (3B). We believe that as we scale up the model size, the advantage of \ours and FiD will keep increasing.

\subsection{Scaling to Hundreds of Demonstrations with \ours}\label{sec:cf_scaling}
    In Section~\ref{sec:cf_dependency}, we show that raising the number of demonstrations $k$ from $16$ to $64$ significantly improves the performance, showcasing the potential of increasing demonstrations. In this section, we further investigate the performance of \ours and FiD when scaling up to \emph{hundreds} of demonstrations \revise{(T5 baseline will run out of GPU memory in these settings)}. As usual, we use {HR$\to$ LR} setting as our testbed. Due to computational limitations, we focus on the \emph{unseen domain} setting where fewer tasks are evaluated. Here are several observations from the scaling curves in Figure~\ref{fig:avn}.

\paragraph{Performance can be further improved when we scale up to hundreds of demonstrations.} By scaling up the number of demonstrations, we can further boost the performance of both \ours and FiD. The results also show that when the number of demonstrations is large, the benefit of \ours is more prominent, suggesting the importance of \revise{enabling the fusion among demonstrations in the encoder}.

\paragraph{Training with larger $k$ can enhance the scalability.} We also explore the influence of the training-time number of demonstrations (train $k$). We observe that for both FiD and \ours, train $k=64$ achieves better performance than train $k=16$ when the test $k\geq 16$. This suggests that a larger training-time number of demonstrations is beneficial for extrapolating to a larger test-time number of demonstrations. It is an interesting future direction to explore how to improve the model's ability to \emph{extrapolate} to larger $k$.

\begin{figure}[t]

\includegraphics[width=\linewidth]{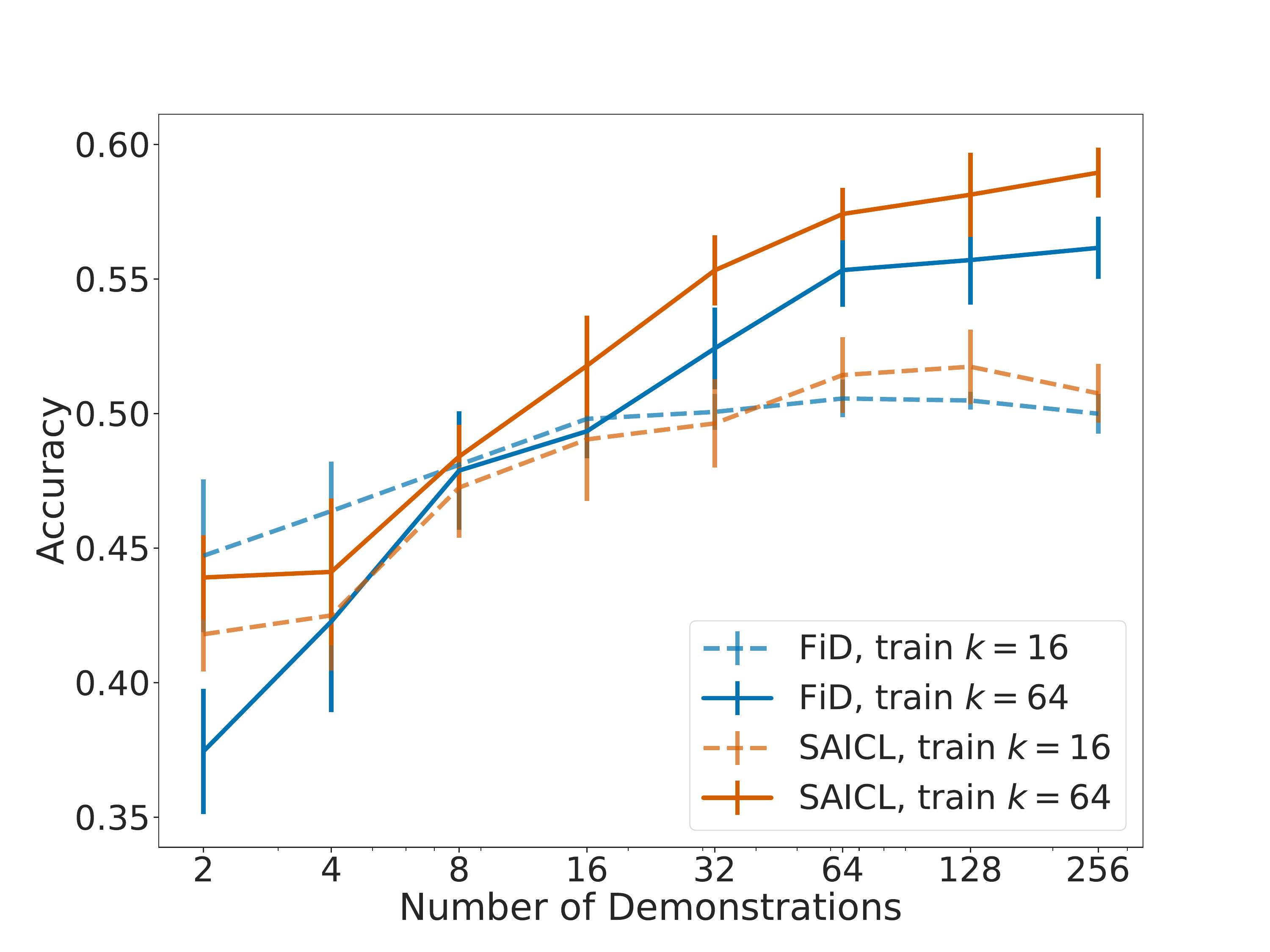}

\caption{Scaling behaviour of FiD and \ours. Generally, the performances of both \ours and FiD improve when the number of demonstrations scales up. \ours achieves significantly better performance when using many demonstrations. A larger train $k$ is beneficial when extrapolating to larger test $k$. The reported performances are tested on {HR$\to$ LR} \emph{unseen domain} setting. %
}
\label{fig:avn}
\end{figure}

\subsection{Combining with Ensemble}\label{sec:cf_ensemble}
    Ensemble methods are commonly used in machine learning. For in-context learning, \citet{min2021noisy, ye2022investigating} show that simply aggregating $k$ predictions based on $k$ single-shot prompts is already an effective approach to achieve extensibility, efficiency, and permutation invariance. However, as shown in \citet{ye2022investigating} and our experiments in Appendix~\ref{sec:ensemble}, the ensemble of single-shot predictions usually under-performs FiD and also saturates fast as $k$ grows. Therefore, in this section, we investigate the hybrid approach of combining ensemble and \emph{few-shot}-prompted models with the hope of achieving better performance at the expense of \emph{losing efficiency and permutation invariance}. Concretely, given $k$ demonstrations, we equally split them into $G$ groups, independently construct the prompts, make predictions over each group, and finally average the logits of all groups to obtain the final prediction. 

We run experiments over combinations of $k\in\{16, 32, \dots, 512\}$ and $ G\in\{1, 2, 4, 8\}$, and compare the baseline T5 model and \ours under {HR$\to$LR} unseen domain setting. We plot the Pareto frontier of the performance versus the inference time in Figure~\ref{fig:pf}. %
As we can see, combining with the ensemble method can further boost the performance of both baseline and \ours. Remarkably, when $k=512$ and $G=8$, \ours achieves $60.8\%$ accuracy, which is $11.8\%$ higher than the $16$-shot \ours baseline, demonstrating the great potential of using more demonstrations. Meanwhile, the performance of the baseline model is also boosted, where the best $59.4\%$ accuracy is achieved when $k=64$ and $G=8$. Generally, for the baseline model, the improvement is more significant when each group is reasonably small (e.g., $8$ demonstrations). And since when each prompt only has a few demonstrations, the efficiency advantage of \ours is less significant (as is shown in Figure~\ref{fig:fs}), we see in the Pareto frontier curve that the baseline dominates when inference is fast. 
\revise{Yet, \ours has a better frontier when the inference time is greater than $800$ ms, i.e., using more demonstrations.}

\begin{figure}[t]
    \includegraphics[width=\linewidth]{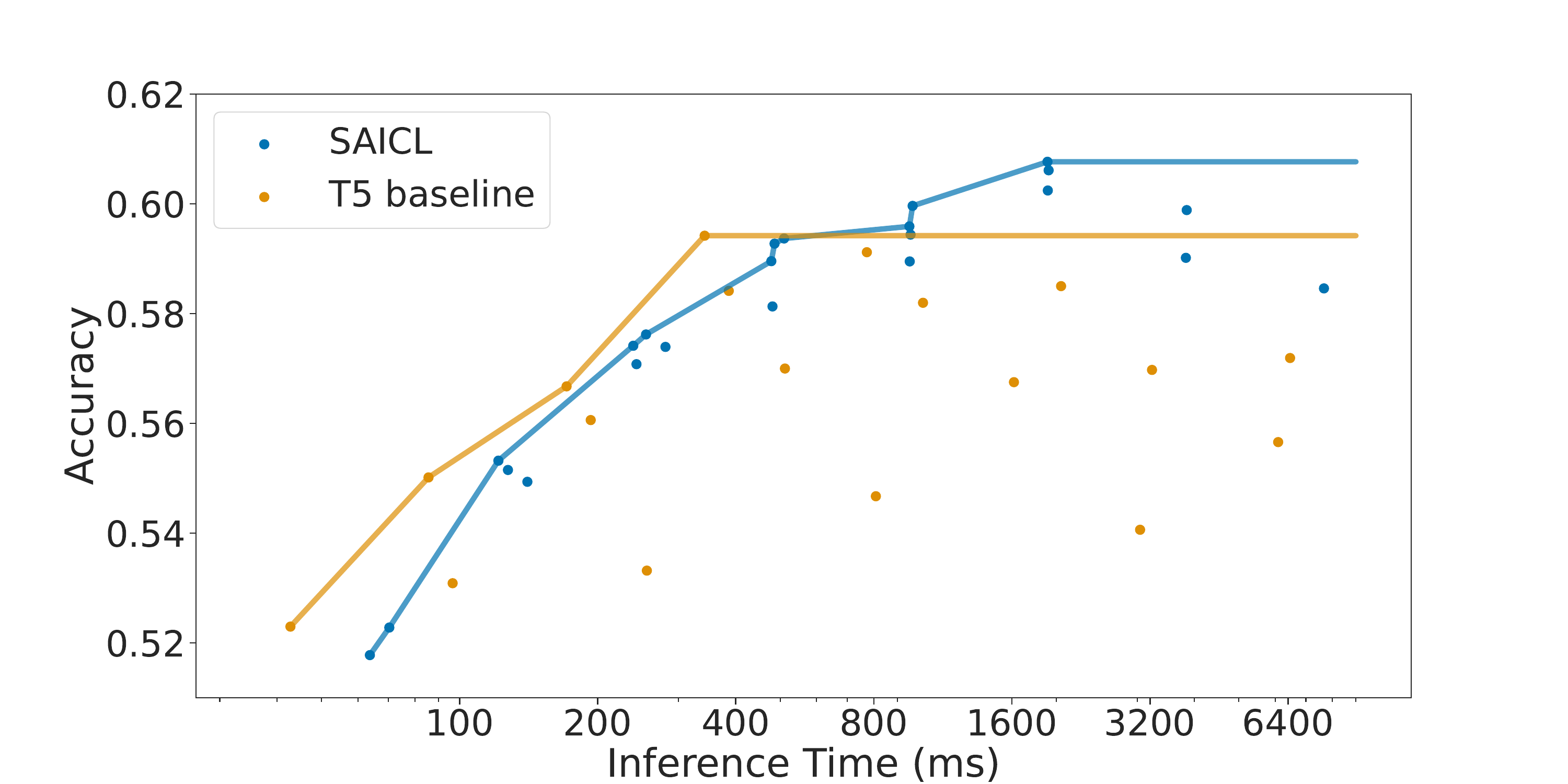}
    \caption{Pareto frontier for accuracy v.s. inference time. We vary the ensemble size in $\{1,2,4,8\}$ and the number of demonstrations in $\{16,32,64,128,256,512\}$ for T5 baseline and \ours. The baseline dominates when the inference time is short. But \ours performs better when using more demonstrations. %
    The accuracies are tested on the {HR$\to$LR} \emph{unseen domain} setting.   
    }
    \label{fig:pf}
\end{figure}

\section{Discussion}\label{sec:discussion}
    This paper takes an initial step toward a better model design for in-context learning, yet many interesting questions remain open. We discuss the important ones below and summarize other discussion questions in Appendix~\ref{sec:faqs}. 

{\paragraph{The applicability of \ours.} In this paper, we use meta-training to fine-tune the pre-trained model due to the modification of the attention mechanism. Other potential ways to apply the idea of \ours include 1) approximating the full attention with \ours, and then directly using pre-trained models without further fine-tuning. This requires careful alignment of the positional encoding; 2) designing unsupervised objectives to pre-train/fine-tune a \ours model. A possible approach is to use an additional retrieval module to find similar sentences as demonstrations and use the language modeling objective on the test sample.}

\paragraph{Extrapolation ability to more demonstrations.} In our experiments of scaling up the number of demonstrations in Section~\ref{sec:cf_scaling}, we observe that, although \ours removes the restriction on the number of demonstrations, the effect of increasing the number of demonstrations $k$ saturates quickly when $k$ is larger than the one used in training time. It is an interesting future direction to explore how to improve the ability of the model to extrapolate to larger $k$.

\paragraph{Combination of different methods.} Following the discussion of extrapolation ability in Section~\ref{sec:cf_scaling}, one possible way to extrapolate to larger $k$ is to combine different fusion methods. The combination with the ensemble methods in Section~\ref{sec:cf_ensemble} is an example of the idea. Moreover, the methods of \citet{hao2022structured, ratner2022parallel} may also be used to further boost the performance when there are many demonstrations available (see our tentative experimental results in Appendix~\ref{sec:ensemble}). An interesting open question will be \emph{how to combine different methods to achieve the best performance, given the number of available demonstrations}.

\paragraph{Understanding in-context learning.} From our experimental results, we see that removing the attention between different demonstrations does not hurt the performance of in-context learning in most settings. Moreover, fusion only in the decoder already suffices to achieve comparable performance to the full attention model. This observation raises the question of \emph{how much \revise{fusion of information} is needed for in-context learning?}. Furthermore, we find different tasks may have different degrees of requirements on the \revise{fusion of information}. For example, classification tasks benefit more from the information exchange between demonstrations, as discussed in Section~\ref{sec:cf_dependency}. %

\section{Conclusion}\label{sec:conclusion}

\revise{In this paper, we seek to improve the architecture design of large language models for in-context learning. We propose \ours, a structured attention mechanism that exploits the prompt structure of in-context learning. The computational and memory complexity of \ours only scales linearly with the number of demonstrations and the model does not have hard-coded restrictions on the number of demonstrations. Furthermore, our method is completely insensitive to the permutations of the demonstrations. }

\revise{We adapt the FiD method from open-book question answering to in-context learning, which also  enjoys extensibility, efficiency, and permutation invariance as our method. We empirically compare our method against the T5 baseline and FiD under the same setting as \citep{min2021metaicl}. Our experiments demonstrate that FiD is a strong baseline, yet \ours outperforms FiD under 18 of 28 settings and is able to match or beat the performance of the T5 baseline ($13$ wins out of $28$ settings). Then, we show that \ours is able to efficiently scale up to hundreds of demonstrations ($6.3$x speed-up when using $128$ demonstrations compared to the baseline) 
and can continuously boost test performance with more demonstrations (e.g., 13.8\% relative performance improvement when we increase $k$ from $16$ to $256$ under HR$\to$LR unseen domain setting).}

\section*{Acknowledgements} Tianle and Kaixuan would like to thank Professor Danqi Chen for wonderful discussions and valuable suggestions.
JDL acknowledges support of the ARO under MURI Award W911NF-11-1-0304,  the Sloan Research Fellowship, NSF CCF 2002272, NSF IIS 2107304,  NSF CIF 2212262, ONR Young Investigator Award, and NSF CAREER Award 2144994.
Mengdi Wang acknowledges the support by NSF grants DMS-1953686, IIS-2107304, CMMI-1653435, ONR grant 1006977, and C3.AI.

\bibliographystyle{icml2023}
\bibliography{in_context_ref}
\appendix
\onecolumn

\clearpage
\section{More Discussion}\label{sec:faqs}
    We summarize several discussion questions about this paper and gather the related discussions in the main paper here for better understanding.

\paragraph{Why use an encoder-decoder model (T5)?} We discuss this in Section~\ref{sec:method}. The core ideas are pasted here: 1) More flexible information exchange. The bidirectional attention mechanism enables free information exchange between different demonstrations. Specifically, in \ours, the information can be passed through the global attention by the test sample. Also, the encoder-decoder separation enables parallel encoding in FiD. 2) Easier to keep the symmetry. It is more natural to use bidirectional attention for encoding demonstrations because there is no canonical order of demonstrations, but the causal mask in decoder-only models induces an order. Also, as mentioned before, the relative positional encoding in T5 enables the permutation invariance in \ours. Yet, we believe that it is easy to adapt our methods to a decoder-only model with modifications on the positional encoding and the attention mask.

\paragraph{Why does permutation invariance matter?} We discuss this in Section~\ref{sec:more_related} and \ref{sec:method}. In summary, this is because 1) The performance is sensitive to the ordering~\citep{zhao2021calibrate,lu2021fantastically}. This makes the results unstable and fixing this requires additional computational costs for searching for a good ordering. 2) The demonstrations are naturally symmetric and the model should fit into this symmetry.

\paragraph{What will happen when scaling up the model size?} We discuss this in Section~\ref{sec:cf_dependency}. We believe it is an interesting future work to extend our methods to larger models. Also, we notice that in the concurrent work \citet{ye2022investigating} (Figure 2), the authors already find that the dominance of FiD compared to ensemble methods is much more significant when the model size increases from T5 large to T5 XL (3B).

\paragraph{Why are there large differences across different settings in Table~\ref{tab:main-result}?} We discuss this in Section~\ref{sec:cf_dependency}. The core ideas are pasted here: Our hypothesis is that solving different tasks requires different forms of information. For example, to solve classification tasks with in-context learning, it is crucial to determine the label space and the format by which a classification problem is converted to natural language. In Figure~\ref{fig:task_type}, we inspect the behaviors of \ours and FiD on different types of tasks under the \emph{test} domain in the HR$\to$LR setting, where all types of tasks are covered by both training and test domains. One observation is that the benefits of \ours are more pronounced on classification tasks. We hypothesize that this is because classification tasks require more subtle knowledge about the format and label space, which benefits from the fusion of demonstrations enabled by \ours in the encoder. This observation also coincides with the finding in \citet{min2022rethinking} that in-context learning relies on the label space, input format, and distribution of input text to work.

\paragraph{Can we pre-train \ours model?} We discuss this in Section~\ref{sec:discussion}. It is a promising direction to explore using unsupervised objectives to pre-train/fine-tune a \ours model. The key challenge here is constructing training samples in the in-context learning format. A possible approach is to use an additional retrieval module to find similar sentences as demonstrations and use the language modeling objective on the test sample.

\paragraph{Why is the speed advantage of \ours less significant when there are fewer demonstrations?} We discuss this in Section~\ref{sec:method}. We hypothesize that this is mainly due to the implementation overhead of \ours. During implementation, we mainly focus on the asymptotic complexity of \ours when scaling the number of demonstrations (As we can see in Figure~\ref{fig:fs}). As a result, the implementation might introduce constant overhead compared to well-optimized full attention. We plan to further investigate the overhead by profiling the computation and trying to optimize the computation kernel of \ours.

\paragraph{Why do other works show that the improvement from scaling demonstrations saturates quickly?} We discuss this in Section~\ref{sec:cf_dependency}. Taking \citet{min2021metaicl} as an example, where the authors show that the performance saturates with only 32 demonstrations. We hypothesize that the difference is because the $1024$ length limitation of the GPT-2 model used in \citet{min2021metaicl} constrains the effective context, and aggressive truncation is needed here. This finding suggests that the potential of boosting performance with more demonstrations might be underestimated because of inappropriate architectural design.
\section{Python-style Pseudo-code of \ours}\label{sec:python_pseudo_code}
\ifdefined\usebigfont
Placeholder
\else
    \begin{minted}{python}
# We heavily use Einstein style notation for tensor operations
from einops import rearrange
# equivalent to torch.einsum
from opt_einsum import contract


def StructuredAttention(query_states, key_states, value_states, mask, segment_length):
    '''
    query_states, key_states, value_states: 
        size: (batch_size, num_heads, num_segment * segment_length, dim)
    mask:
        size: (batch_size, 1, 1, num_segment * segment_length)
    segment_length: 
        max length for each demonstration/test input

    '''

    # split into blocks
    # (batch_size, num_heads, num_seg, segment_length, dim)
    query_states = rearrange(
        query_states, 'b h (s t) d -> b h s t d', t=segment_length)
    key_states = rearrange(
        key_states, 'b h (s t) d -> b h s t d', t=segment_length)
    value_states = rearrange(
        value_states, 'b h (s t) d -> b h s t d', t=segment_length)
    # (batch_size, 1, num_seg, 1, key_length)
    block_mask = rearrange(mask, 'b h t (s r) -> b h s t r', r=segment_length)

    # the diagonal part
    # shape (batch_size, num_heads, num_seg, segment_length, segment_length)
    block_diag = contract(
        'b h s t d, b h s r d -> b h s t r', query_states, key_states)

    # calculate position bias for each block
    # this function follows the original T5 relative position bias
    position_bias_diag = calculate_position_bias(
        segment_length, segment_length)
    # (batch_size, num_heads, 1, segment_length, segment_length)
    block_diag_bias = rearrange(position_bias_diag, 'b h t r -> b h 1 t r')
    # add bias to diagonal part
    block_diag += block_diag_bias + block_mask

    # last block column, attention from all demonstrations to the test sample
    block_global_key = contract(
        'b h s t d, b h r d -> b h s t r', query_states[:, :, :-1], key_states[:, :, -1])
    block_global_key += block_mask[:, :, -1, None]
    # last row, attention from the test sample to all demonstrations
    block_global_query = contract(
        'b h t d, b h s r d -> b h s t r', query_states[:, :, -1], key_states[:, :, :-1])
    block_global_query += block_mask[:, :, :-1]

    # merge block diagonal and last column for computing softmax
    block_global_key_cat = torch.cat(
        [block_diag[:, :, :-1], block_global_key], dim=-1)
    block_global_key_cat = nn.functional.softmax(
        block_global_key_cat.float(), dim=-1).type_as(block_global_key_cat)
    # merge last row for computing softmax
    block_global_query_cat = torch.cat(
        [block_global_query, block_diag[:, :, -1, None]], dim=2)
    block_global_query_cat = rearrange(
        block_global_query_cat, 'b h s t r -> b h t (s r)')
    block_global_query_cat = nn.functional.softmax(
        block_global_query_cat.float(), dim=-1).type_as(block_global_query_cat)

    # dropout
    block_global_key_cat = nn.functional.dropout(
        block_global_key_cat, p=self.dropout, training=self.training
    )
    block_global_query_cat = nn.functional.dropout(
        block_global_query_cat, p=self.dropout, training=self.training
    )

    # split back for computing block matrix multiplication
    output_key_diag = contract('b h s t r, b h s r d -> b h s t d',
                block_global_key_cat[..., :segment_length], value_states[:, :, :-1])
    output_key_global = contract('b h s t r, b h r d -> b h s t d',
                    block_global_key_cat[..., segment_length:], value_states[:, :, -1])
    output_query_global = contract('b h r l, b h l d -> b h r d', block_global_query_cat,
                    rearrange(value_states, 'b h s t d -> b h (s t) d'))
    output_query_global = rearrange(
        output_query_global, 'b h r d -> b h 1 r d')
    output = torch.cat([output_key_diag+output_key_global,
                       output_query_global], dim=2)

    attn_output = rearrange(
        output, 'b h s t d -> b (s t) (h d)', t=segment_length)

    return attn_output

\end{minted}
\fi
\section{Detailed Experimental Settings}\label{sec:detail_setting}
    \paragraph{Data processing.} We follow the MetaICL codebase to process the data. The input in the in-context learning setting consists of several demonstrations and the test sample. We first choose a sample-wise max length ($256$ for T5 baseline models and $\min(256, \text{max length in the same task})$ for \ours and FiD), then truncate each demonstration $(y_i, \xbf_i)$ and $y_\mathrm{test}$ when the length exceeds the max length and pad to max length when using \ours and FiD. We pack as many demonstrations as the number of demonstrations is at most $k$, and the total length of the input is at most $64k$ as the final input. This way, the final number of demonstrations will range from $k/4$ to $k$, and for most tasks, it will be $\approx k$.

\paragraph{Training details.} We use Adafactor optimizer~\citep{shazeer2018adafactor} with learning rate $1e-4$ (following previous works of fine-tuning T5~\citep{ye2022guess}) and batch size $4$. With the linear learning rate schedule, the learning rate first linearly increases from $0$ to $1e-4$ for $10\%$ steps and then linearly decreases to $0$. The total optimization steps we use is $25600$.
\section{Additional Experimental Results}\label{sec:ensemble}
    We summarize an overview of different methods in Table~\ref{tab:method}. Specifically, when the prompt contains multi-shot demonstrations, in the encoder, we can choose between sparse attention (i.e., \ours) and full attention. When we need to fuse the information from multiple prompts, we can choose between (1) feeding them independently through the decoder and averaging the logits, and (2) concatenating all the processed tokens and feeding them once to the decoder. We use \emph{Group-FiD} to refer to the case when we construct multiple prompts, each with multiple demonstrations, and concatenate all the encoded tokens as the input to the decoder. 

\begin{table*}[h] 
    \centering
    \begin{tabular}{lcccc}
        \toprule
           Method & Encoder Attention  & Fusion Scheme & Permutation  Invariance? & Linear Complexity? \\ 
        \midrule
        -  & Single-shot & Single-prompt & - & -  \\
        Baseline & Full & Single-prompt & No & No \\
        \ours & Sparse & Single-prompt & Yes& Yes\\
         Ensemble $k=1$ & Single-shot &Average & Yes & Yes\\
         Baseline Ensemble & Full & Average & No & No\\
         \ours Ensemble & Sparse & Average & Yes& Yes\\
         FiD & Single-shot & Concat & Yes& Yes\\
          Baseline + Group-FiD & Full &  Concat & No& No \\
         \ours + Group-FiD & Sparse & Concat & Yes& Yes\\
        \bottomrule
    \end{tabular}%
    \caption{Comparison of different methods. \textit{Single-shot} means each prompt only contains one demonstration.  \textit{Single-prompt} means we only construct one prompt, which may contain multiple demonstrations.  \textit{Sparse} means we use \ours in the encoder attention, and \textit{Full} means we use the standard full attention. \textit{Average} means we average the logits of all the predictions from different prompts, and \textit{Concat} means we concatenate all the encoded tokens from different prompts as the input to the decoder. 
    }\label{tab:method}
\end{table*}

\subsection{Comparison among Single-shot-prompted Methods  }

We test the performance of ensembling T5 baseline models where each instance only contains $1$ demonstration and compare the results with the Fusion-in-Decoder method. 
For both methods, the $k$ single-shot prompts are processed independently through the encoder. The difference is that, in the fusion-in-decoder method, the processed tokens are concatenated for computing the logit in the decoder, while for the direct ensemble method, the processed tokens are used for computing the logit in the decoder independently and then the logits are averaged.

The results are reported in Figure~\ref{fig:a1}. We see that the ensemble of single-shot predictions performs worse than FiD method and saturates when $k \geq 8$.
All the results are tested using the corresponding models with train $k=64$ on HR$\to$LR \emph{unseen domain} setting.

\begin{figure*}[htbp]
    \centering
    \subfigure[]{
        \label{fig:a1}
        \includegraphics[width=0.7\linewidth]{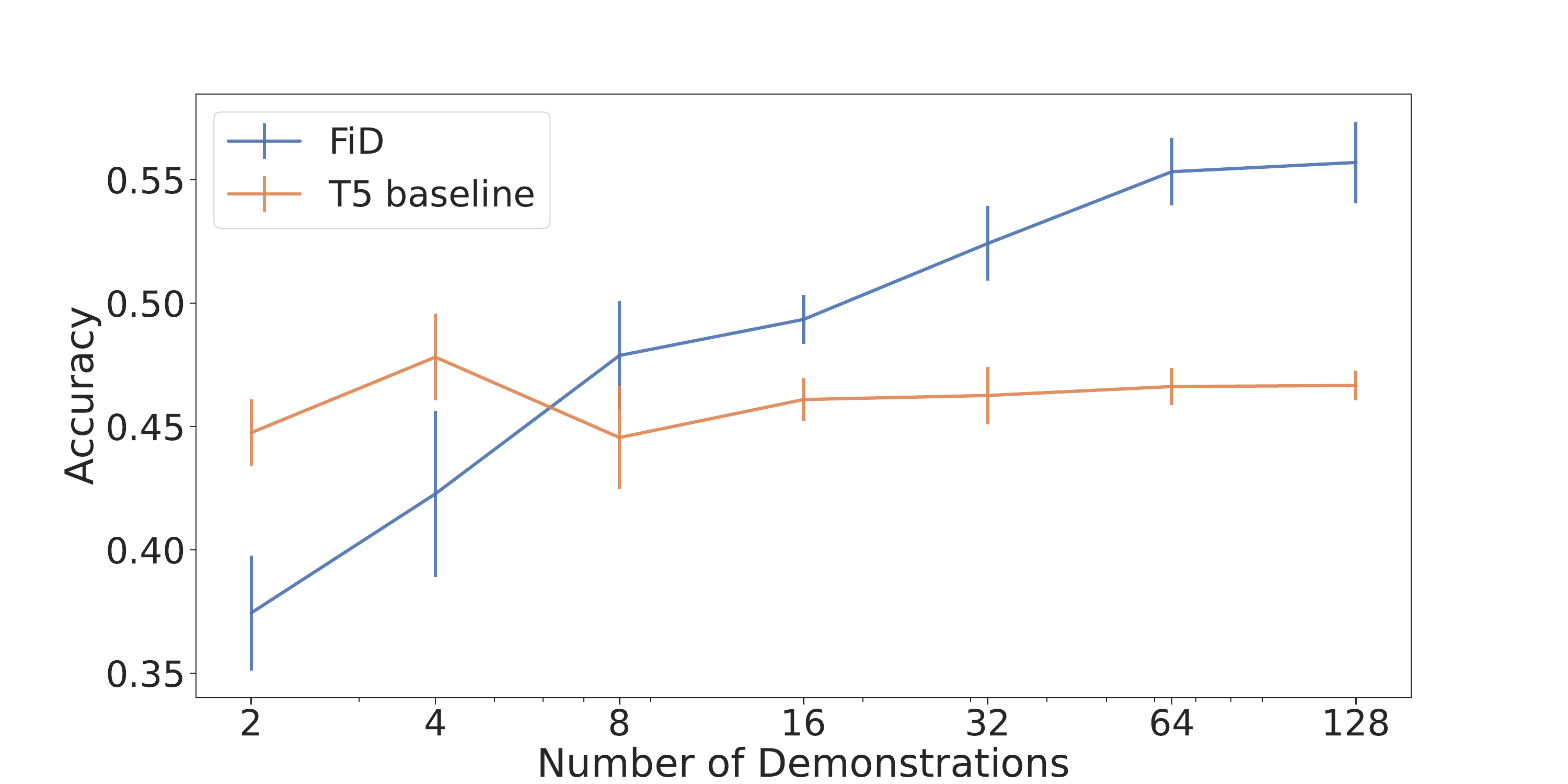}
    }\\
    \subfigure[]{
        \label{fig:a2}
        \includegraphics[width=0.7\linewidth]{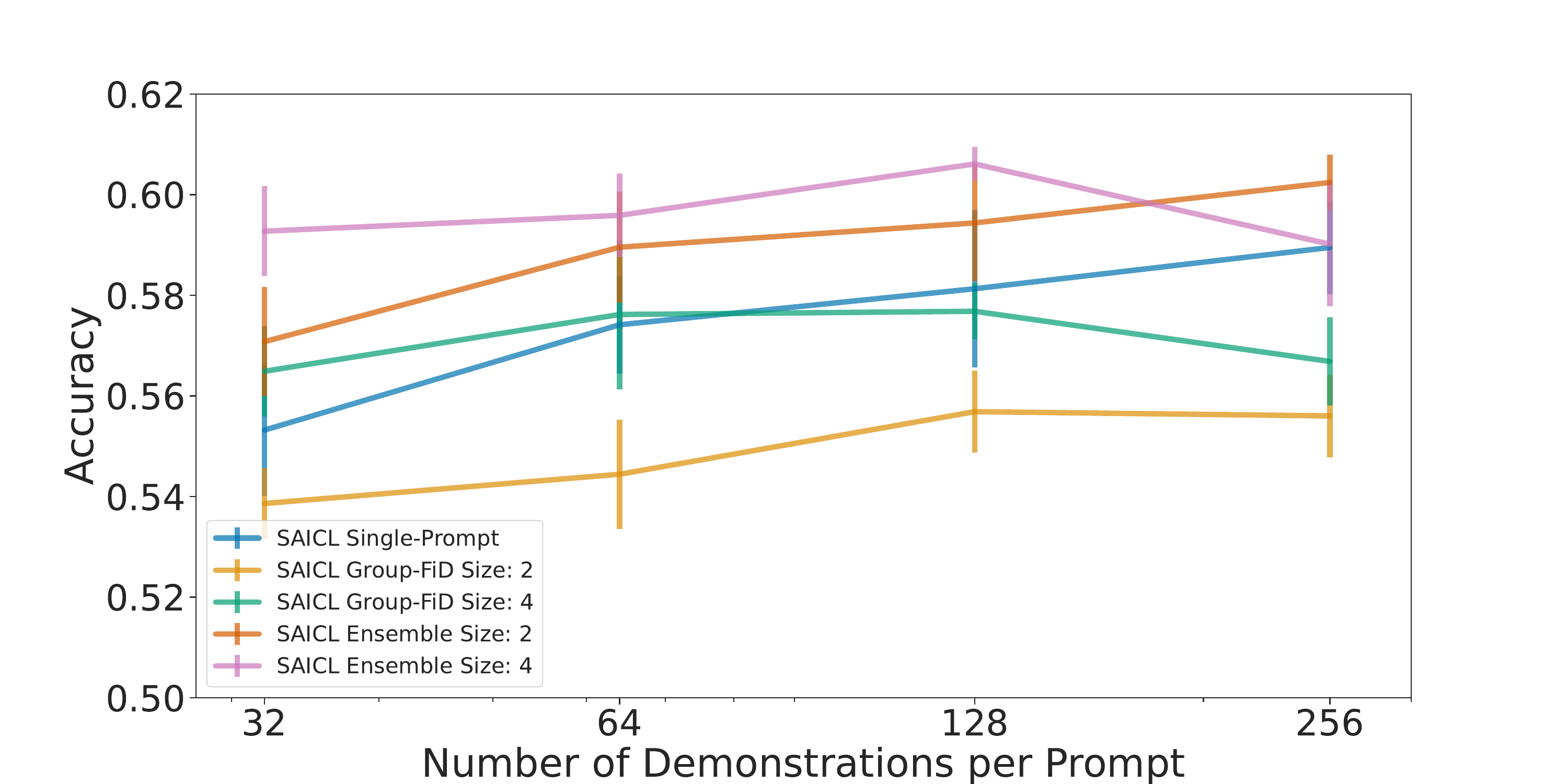}}\\
    \subfigure[]{
        \label{fig:a3}
        \includegraphics[width=0.7\linewidth]{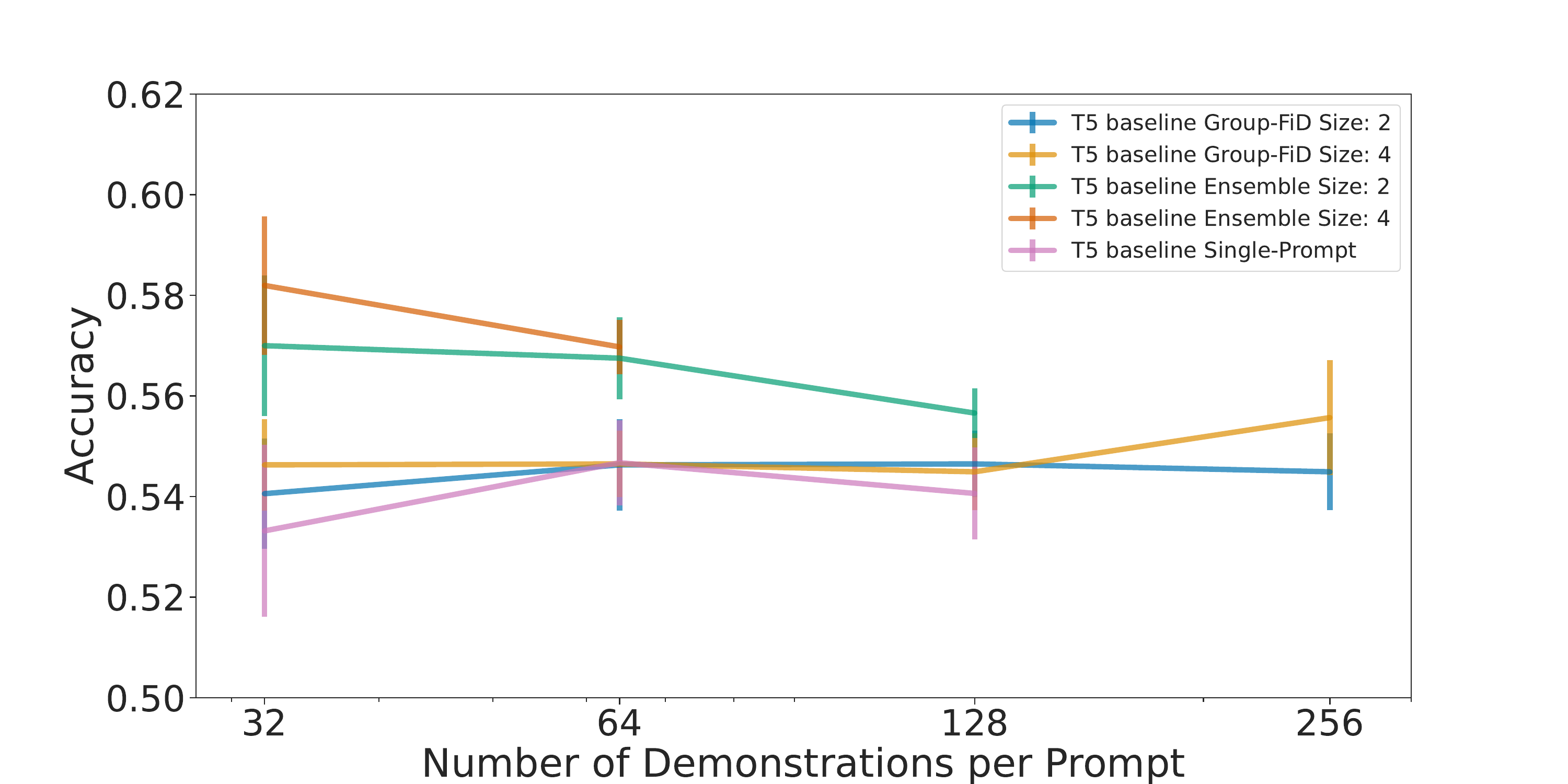}}
    \caption{ \textbf{(a)}: Comparison between Ensemble and FiD. We see that direct ensembling the predictions usually perform worse than the FiD method, and saturate quickly when $k\geq 8$. \textbf{(b)}: Comparison between Ensemble and Group-FiD for \ours. We see that Ensemble performs better than Group-FiD. \textbf{(c)}: Comparison between Ensemble and Group-FiD for T5 baseline. We see that Ensemble performs better than Group-FiD. All the results are tested using the corresponding models with train $k=64$ on HR$\to$LR \emph{unseen domain} setting.} 
\end{figure*}

\subsection{Comparison between Group-FiD and Ensemble}

In the subsection, we test \emph{Group-FiD} method,  which resembles the approach proposed in \citep{hao2022structured, ratner2022parallel}. In Group-FiD, we divide the demonstrations equally among $G$ groups, where each group may possesses multiple demonstrations. We pass these groups independently through the encoder and then concatenate the processed tokens and send them into the decoder. The method can be combined with both \ours and the full attention baseline. 

We compare Group-FiD with ensemble and report the results for \ours in Figure~\ref{fig:a2} and T5 baseline in Figure~\ref{fig:a3}. We see that for both \ours and the baseline, Group-FiD scheme falls behind the ensemble scheme.
All the results are tested using the corresponding models with train $k=64$ on HR$\to$LR \emph{unseen domain} setting.

\section{Other Related Work}\label{sec:more_related}
    \paragraph{(Few-shot) fine-tuning.} The traditional way of adapting a pre-trained language model to new tasks is to fine-tune the model. Recent works on instruction tuning~\citep{sanh2021multitask,wei2021finetuned} show that fine-tuning language models on a collection of datasets described via instructions—substantially improves \emph{zero-shot} performance on unseen tasks. Yet, these methods require a diverse set of datasets with a large amount of data, and the training cost is usually high because the whole model has to be updated. Several methods seek to solve the efficiency problem and fine-tune the model in a parameter-efficient way. This is achieved by only tuning the prompt~\citep{lester2021power,li2021prefix,gao2021making,liu2021p,an2022input,chen2022adaprompt}, tuning a subset of the parameters~\citep{zaken2022bitfit,sung2021training,guo2021parameter}, or adding a few additional tunable parameters and fixing the original model~\citep{rebuffi2017learning,houlsby2019parameter,henderson2021compacter,liu2022few,bapna2019simple,hu2021lora}. These methods can largely reduce the computational cost of fine-tuning, but they still need at least thousands of updates and thousands of samples. In comparison, in-context learning only involves one forward pass and requires fewer samples. With our methods, the forward complexity only linearly scales to the number of demonstrations, which is more efficient fine-tuning. %

\end{document}